%% file: 0_main.tex
\documentclass[conference,letterpaper,10pt]{IEEEtran}
\IEEEoverridecommandlockouts

\input{01_config}

\begin{document}

\AddToHookNext{shipout/foreground}{%
\begin{tikzpicture}[overlay, remember picture]
    \node at ([yshift=-1cm]current page.north) {
        \normalsize\textcolor{gray}{This paper has been accepted for publication at the}
    };
    \node at ([yshift=-1.5cm]current page.north) {
        \normalsize\textcolor{gray}{IEEE International Conference on Omni-Layer Intelligent Systems (COINS), Bologna, Italy, 2026}
    };
\end{tikzpicture}%
}

\title{LiZAD: A Lightweight Zero-Shot Anomaly Detection Framework for Industrial Manufacturing}

\author{
\IEEEauthorblockN{
Uzair Khan\IEEEauthorrefmark{1},
Luigi Capogrosso\IEEEauthorrefmark{2},
Muhammad Aqeel\IEEEauthorrefmark{1},
Francesco Setti\IEEEauthorrefmark{1},\\
Michele Magno\IEEEauthorrefmark{3}\IEEEauthorrefmark{2},
Marco Cristani\IEEEauthorrefmark{1}}
\IEEEauthorblockA{
\IEEEauthorrefmark{1}University of Verona,
\IEEEauthorrefmark{2}Interdisciplinary Transformation University of Austria,
\IEEEauthorrefmark{3}ETH Zurich}
}

\IEEEpubid{\makebox[\columnwidth]{
979-8-3195-0489-0/26/\$31.00 ©2026 IEEE\hfill}
\hspace{\columnsep}\makebox[\columnwidth]{ }}
\IEEEpubidadjcol

\maketitle

\bstctlcite{IEEEexample:BSTcontrol}

%%%%%%%%% ABSTRACT.
\begin{abstract}
In modern high-throughput industrial production lines, product configurations and visual characteristics frequently change, making it impractical to collect and annotate data for every new scenario.
This dynamic setting makes Zero-Shot Anomaly Detection (ZSAD) particularly suitable, as it enables defect detection without requiring training on target-specific samples.
Although recent ZSAD approaches show promising results, they are computationally intensive and thus unsuitable for deployment on resource-constrained devices.
We propose \ours{}: a lightweight framework designed for real-time ZSAD specifically tailored for use on edge devices.
The proposed approach pairs the dense and spatially aware visual features of DINOv3, crucial for precise pixel-level localization, with the highly computationally efficient text embeddings of MobileCLIP2.
These features are then mapped into a shared latent space via low-memory trainable projection heads.
Compared to six state-of-the-art ZSAD models, \ours{} achieves an average memory reduction of 61.5\%, a parameter reduction of 74.6\%, and a speedup of 3.02$\times$ in terms of latency.
Despite substantial reductions in computational and memory costs, our approach maintains competitive anomaly detection performance, dropping the average P-AUROC by just 6.4\% relative to the best state-of-the-art model across the VisA, BTAD, MPDD, and MVTec-AD datasets.
Finally, it is successfully deployed on the NVIDIA Jetson NX and Jetson AGX edge devices and tested on the real production line of the Industrial Computer Engineering Laboratory (ICE Lab) at the University of Verona.
The code is available at \url{https://github.com/intelligolabs/LiZAD}.
\end{abstract}

%%%%%%%%% BODY TEXT.
\input{src/1_intro}
\input{src/2_related}
\input{src/3_method}
\input{src/4_experiments}
\input{src/5_conclusions}

%%%%%%%%% BIBLIOGRAPHY.
\bibliographystyle{IEEEtran}
\bibliography{bibliography}

\end{document}

%% file: 01_config.tex
\usepackage[T1]{fontenc}
\usepackage[english]{babel}
\usepackage{textcomp}
\usepackage[table, pdftex, dvipsnames]{xcolor}
\usepackage{xspace}
\usepackage{graphicx}
\usepackage{amsfonts}
\usepackage{amsmath}
\usepackage{amssymb}
\usepackage{booktabs}
\usepackage[acronym]{glossaries}
\usepackage[capitalize,noabbrev]{cleveref} % noabbrev
\usepackage{cite}
\usepackage{tikz}
\usepackage{multirow}
\usepackage{multicol}
\usepackage{colortbl}
\usepackage{siunitx}
\usepackage{algpseudocode}
\usepackage{algorithm}
\usepackage{url}
\usepackage[utf8]{inputenc}
\usepackage[most]{tcolorbox}
\usepackage{subcaption}
\glsdisablehyper
\AtBeginDocument{}
\tcbset{
    colback=gray!5,
    colframe=gray!40,
    left=6pt,
    right=6pt,
    top=6pt,
    bottom=6pt,
    boxsep=0pt,
}

\makeatletter
\DeclareRobustCommand\onedot{\futurelet\@let@token\@onedot}
\def\@onedot{\ifx\@let@token.\else.\null\fi\xspace}
\def\eg{\emph{e.g}\onedot} 
\def\ie{\emph{i.e}\onedot}

\makeatother

\newcommand{\ours}{LiZAD\xspace}
\newcommand{\medal}[3]{%
    \tikz[baseline=(char.base)]{%
        \node[rounded corners=2pt,fill=#1,draw=#2,inner sep=1.5pt] (char) {#3};%
    }%
}
\newcommand{\bm}[2]{%
    \ifnum#1=1
        \medal{gold}{gold_dark}{\textbf{#2}}%
    \else\ifnum#1=2
        \medal{silver}{silver_dark}{#2}%
    \else\ifnum#1=3
        \medal{bronze}{bronze_dark}{#2}%
    \else
        #2%
    \fi\fi\fi
}

\definecolor{red}         {HTML}{b7211f}
\definecolor{gold}        {HTML}{FBF2D2}
\definecolor{green}       {HTML}{147546}
\definecolor{silver}      {HTML}{DDDDDD}
\definecolor{bronze}      {HTML}{EED2B8}
\definecolor{gold_dark}   {HTML}{D9AE13}
\definecolor{silver_dark} {HTML}{909090}
\definecolor{bronze_dark} {HTML}{9A5F26}

%% file: src/1_intro.tex
\section{Introduction} \label{sec:intro}

In industrial manufacturing, real-time detection of defective products is critical to ensuring high quality standards, reducing waste, and optimizing production efficiency \cite{Liu2024}.
The ability to detect defective products early prevents costly recalls and improves customer satisfaction \cite{Aqeel2025}.
Traditional supervised anomaly detection has shown strong performance in this task \cite{Khan2025}.
However, its effectiveness is highly dependent on the quantity and diversity of positive and negative labeled samples available.
Furthermore, acquiring such data for every newly introduced product is costly, time-consuming, and often infeasible in practice \cite{Capogrosso2024a, Girella2024}.
Thus, researchers have focused on unsupervised anomaly detection, where models are trained exclusively on negative samples and learn to identify deviations from the learned distribution \cite{Roth2022, Cao2023}.
Although this approach alleviates the need for defective labeled samples, it still requires an operator to carefully curate the training dataset to ensure that no anomalous samples are present \cite{Aqeel2025}.

\begin{figure}[t!]
    \centering
    \includegraphics[width=\linewidth]{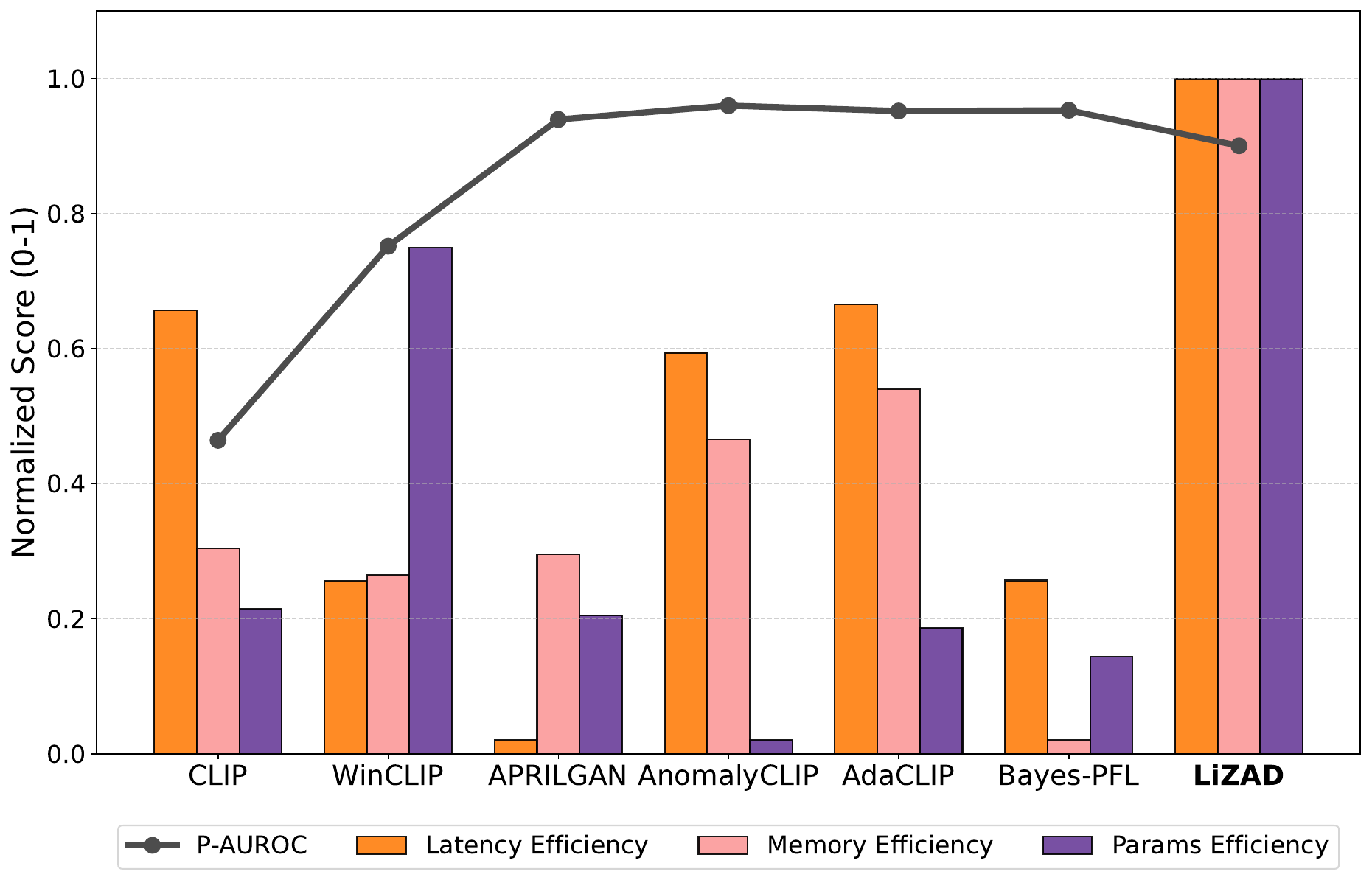}
    \caption{The comparative analysis of \ours{} against six ZSAD models.
    All metrics are normalized to a [0, 1] scale, where 1.0 represents the best achievable performance.
    }
    \label{fig:Teaser}
\end{figure}

These limitations have driven growing interest in Zero-Shot Anomaly Detection (ZSAD), which aims to detect anomalies in previously unseen categories without using any training images of those categories \cite{Jeong2023}.
Despite their promise, existing ZSAD methods still face significant practical limitations in real-world production environments.
First, many current approaches are built on large vision-language backbones and additional heavy components, resulting in high computational and memory requirements, as shown in \cref{fig:Teaser}.
Thus, such systems are difficult to deploy on resource-constrained devices, edge platforms, or cost-sensitive inspection lines.
Second, their inference speed is often too slow for modern industrial workflows, where inspection must keep pace with high-throughput production \cite{Stropeni2026}.
Therefore, beyond detection accuracy, deployability and efficiency are equally important requirements for such industrial anomaly detection systems.

To address the limitations of current models, we propose \ours{}, a lightweight framework designed for real-time ZSAD on resource-constrained edge devices.
It combines DINOv3 \cite{Simeoni2025} features, which provide strong spatial representations for fine-grained anomaly localization, with MobileCLIP2 \cite{Faghri2025} text embeddings, which enable low-cost semantic guidance.
\ours{} achieves state-of-the-art computational efficiency, reducing the average inference latency by 69.9\%.
Furthermore, it reduces resource barriers for edge deployment by cutting the memory requirements by 61.5\% (to 1.01~GB) and the total parameter count by 74.6\% (to 97.44~M).
Remarkably, these efficiency gains incur only a minimal localization trade-off; \ours{} maintains a highly competitive average pixel-wise AUROC of 88.05\%, representing only a 6.4 percentage point drop compared to the absolute best-performing baselines.
We deployed \ours{} on the NVIDIA Jetson NX and Jetson AGX edge devices and qualitatively tested it on the MMS dataset \cite{Bonazzi2026} and on the products of the real production line of the Industrial Computer Engineering Laboratory (ICE Lab) at the University of Verona \cite{Cunico2024}.

%% file: src/2_related.tex
\section{Related Works} \label{sec:related}

%%%%%%%%%%%%%%%%%%%%
\subsection{Industrial Anomaly Detection}

%%%%%%%%%%
\emph{\textbf{Reconstruction-based methods}} detect anomalies by comparing an input image with its reconstruction, assuming that models trained only on normal data will fail to recreate abnormal regions.
Approaches range from early Autoencoders and Variational Autoencoders to more recent inpainting strategies that randomly mask image patches \cite{Zavrtanik2021a, Ristea2022}.
However, their success heavily depends on the overall quality of the reconstruction.
If a decoder is overly expressive or anomalies share common patterns (\eg{}, local edges) with normal data, defects might be accurately reconstructed, resulting in small residual errors that do not distinguish anomalies \cite{Zavrtanik2021a}.

%%%%%%%%%%
\emph{\textbf{Embedding-based methods}} leverage pre-trained networks to extract features and establish a model of normality.
Strategies include memory bank methods \cite{Roth2022, Bae2023} using metric learning, one-class classifiers \cite{Reiss2021} that tightly cluster normal data in a latent space, and distribution approaches \cite{Defard2021} that apply pixel-wise Gaussian models.
Additionally, Normalizing Flow methods \cite{Yu2021, Rudolph2021, Gudovskiy2022} learn to transform normal features into a standard probability space to handle complex data.
Despite their high accuracy, the computational burden of searching memory banks \cite{Roth2022} or computing inverse covariances \cite{Defard2021} severely limits their real-time feasibility on edge devices.

%%%%%%%%%%
\emph{\textbf{Synthesis-based methods}} utilize data augmentation to generate ``fake'' defects on normal images, teaching the model to recognize visual deviations \cite{Capogrosso2024a, Capogrosso2024c, Girella2024}.
Pioneering approaches like CutPaste \cite{Li2021} create anomalies by cutting and pasting image patches, while DRAEM \cite{Zavrtanik2021b} simulates realistic textures using Perlin noise to train defect repair networks.
Recently, SimpleNet \cite{Liu2023} demonstrated high accuracy by training a simple discriminator to differentiate between normal features and synthetic noise.
Building on this foundation, GLASS \cite{Chen2024} further improves weak defect detection by integrating image-space corruptions with feature-space perturbations.

%%%%%%%%%%
\emph{\textbf{Gaps in the literature.}}
Despite their effectiveness, these methods are primarily designed for unsupervised or self-supervised industrial anomaly detection, where at least target-domain normal data are available for model fitting or memory construction.

%%%%%%%%%%%%%%%%%%%%
\subsection{Zero-Shot Anomaly Detection} \label{ssec:zsad}
Recent ZSAD methods increasingly build on pretrained Vision-Language Models to transfer generic visual priors to unseen industrial categories without target-specific training data.
WinCLIP \cite{Jeong2023} extends CLIP \cite{Radford2021} with compositional prompt ensembling and window-based feature matching to perform zero-shot anomaly classification and segmentation.
APRILGAN \cite{Chen2023} further demonstrates CLIP-based knowledge transfer for ZSAD by adding linear adapter layers that map CLIP visual features into the joint image--text embedding space.
AnomalyCLIP \cite{Zhou2023} improves cross-domain generalization by learning object-agnostic prompts that emphasize generic notions of normality and abnormality rather than category-specific semantics.
AdaCLIP \cite{Cao2024} further adapts CLIP by combining shared and instance-aware semantic cues through hybrid learnable prompts, thus enhancing robustness across unseen domains.
Bayes-PFL \cite{Qu2025} models the prompt space from a Bayesian perspective, enabling uncertainty-aware prompt adaptation and more robust anomaly prediction.

%%%%%%%%%%
\emph{\textbf{Gaps in the literature.}}
Existing zero-shot models require too much memory and processing time to be practical on actual industrial production lines.
To the best of our knowledge, \ours{} is the first fast, lightweight, and accurate ZSAD framework built specifically for real-world deployment.

%%%%%%%%%%%%%%%%%%%%
\subsection{Embedded Industrial Anomaly Detection}
Most existing anomaly detection approaches, regardless of the learning paradigm, assume the availability of abundant computational resources and do not account for the strict real-time processing requirements of actual manufacturing environments \cite{Stropeni2026}.
Although recent efforts have introduced efficiency-oriented adaptations, such as PaSTe \cite{Barusco2025}, or developed memory-compressed variants, such as PatchCore-Lite and PaDiM-Lite \cite{Stropeni2026}, the balance between high-speed inference and accurate detection of complex unseen anomalies remains a significant challenge. 
Furthermore, the use of vanilla vision foundation models for embedded anomaly detection is severely constrained by their massive parameter counts; this has led to the development of lightweight SAM-guided architectures such as STLM \cite{Li2024} and KairosAD \cite{Khan2025}.
Finally, the transition towards true ultra-low-power anomaly detection is still largely unexplored.
Bridging this gap requires extreme architectural compression and hardware-aware quantization \cite{Capogrosso2024b}.
This is demonstrated by models such as TinyGLASS \cite{Bonazzi2026}, which adapts self-supervised architectures for direct execution on intelligent vision sensors.

%%%%%%%%%%
\emph{\textbf{Gaps in the literature.}}
Although recent advancements have successfully reduced the computational footprint of visual anomaly detection models, none of these embedded approaches operate in a zero-shot setting.

%% file: src/3_method.tex
\section{Methodology} \label{sec:method}

\begin{figure*}[t!]
    \centering
    \includegraphics[width=\linewidth]{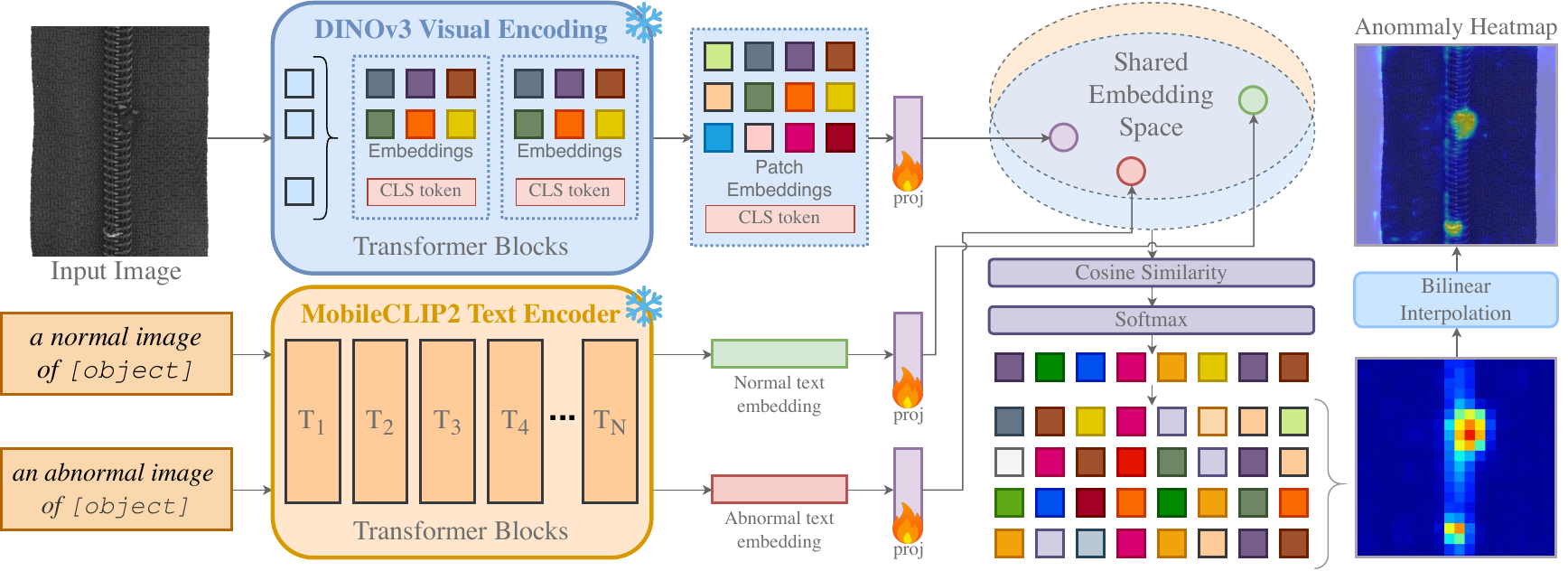}
    \caption{Overview of \ours{} architecture.}
    \label{fig:LiZAD}
\end{figure*}

%%%%%%%%%%%%%%%%%%%%
\subsection{Problem Formulation}
ZSAD aims to detect anomalies in previously unseen object categories without access to target-domain training data.
Let $D_a$ and $D_e$ denote the auxiliary training and evaluation datasets, respectively, and let $\mathcal{C}_a$ and $\mathcal{C}_e$ denote their corresponding category sets.
In the zero-shot setting, the training and evaluation categories are disjoint, \ie{}, $\mathcal{C}_a \cap \mathcal{C}_e = \varnothing$.
In this setting, a model trained on $D_a$ must generalize to images from $D_e$ without observing their categories during training.
Thus, given an evaluation image $x \in D_e$, \ours{} predicts anomalies at the pixel level through an anomaly map $\hat{Y} \in \{0,1\}^{H \times W}$, where higher values indicate a higher likelihood of anomaly.

%%%%%%%%%%%%%%%%%%%%
\subsection{Overview of LiZAD}
Existing ZSAD methods are commonly based on CLIP-style models (details in \cref{ssec:zsad}).
Although such models are effective for global image-text matching, their visual representations are primarily optimized for semantic alignment and are often less suitable for fine-grained anomaly localization.
However, in industrial inspection, defects are frequently subtle, spatially localized, and difficult to identify from global representations alone.

To address this issue, we adopt DINOv3 as a visual backbone \cite{Simeoni2025}, as shown in \cref{fig:LiZAD}.
To ensure that our framework remains computationally efficient without sacrificing representational capacity, we adopt the compact ViT-S/16 variant that uses a Gram anchoring regularization during its self-supervised pre-training.
This explicitly preserves strong dense patch-level representations and prevents feature degradation, making it well-suited for fine-grained anomaly localization.

For the text branch, we adopt MobileCLIP2 \cite{Faghri2025}, an architecture explicitly engineered for fast and low-latency execution on edge hardware through a highly optimized convolution-transformer design.
Furthermore, it overcomes the typical representational drop of lightweight networks through multi-modal reinforced training, acting as a sophisticated knowledge distillation process from massive teacher models.
This ensures that our framework extracts highly expressive text embeddings with minimal memory footprint and computational overhead at inference time.

During inference, we use the same normal and abnormal prompt templates as during training for the target object category.
The input image is first encoded into patch-level features, which are then projected into the shared embedding space.
The coarse patch-level anomaly map is computed using the cosine similarity between the projected visual patch features and the normal/abnormal text embeddings.
Finally, the map is upsampled to the input resolution to obtain the final dense anomaly heatmap.

%%%%%%%%%%%%%%%%%%%%
\subsection{Shared Embedding Alignment} \label{ssec:embedding-lignment}
Given an input image, \ours{} extracts a global visual token and patch-level embeddings using a frozen DINOv3 encoder, while normal and abnormal text prompts are encoded by a frozen MobileCLIP2 text encoder.

Formally, let $f_v$ and $f_t$ denote the frozen DINOv3 image encoder and the frozen MobileCLIP2 text encoder, respectively.
For an input image $x$, the visual encoder produces a global token $c \in \mathbb{R}^{D_v}$ and a set of patch embeddings $\{p_i\}_{i=1}^{N}$, where $p_i \in \mathbb{R}^{D_v}$ and $N$ is the number of spatial patches.
For normal and anomalous text prompts $q_n$ and $q_a$, the text encoder outputs text embeddings $t_n \in \mathbb{R}^{D_t}$ and $t_a \in \mathbb{R}^{D_t}$.

Because $D_v \neq D_t$ in general and the two encoders are not jointly trained, a direct comparison of similarity is not meaningful.
We therefore learn lightweight linear projections to map the global token, patch embeddings, and text embedding into a common $D$-dimensional space:
\begin{equation}
    \hat{c} = \mathrm{proj}_{c}(c)\;, \quad 
    \hat{p}_i = \mathrm{proj}_{p}(p_i)\;, 
\end{equation}
where $\mathrm{proj}_{c}(\cdot)$ and $\mathrm{proj}_{p}(\cdot)$ are learnable projection heads.

Following standard text-guided anomaly detection protocols \cite{Zhou2023, Ma2025, Qu2025}, we construct normal and abnormal prompts using the template \texttt{``an image of a \{state\} \{object\}''} where \texttt{state} is \texttt{normal} or \texttt{abnormal}, and \texttt{object} denotes the object category.
The text embeddings are obtained as:
\begin{equation}
    t_n = f_t(q_n)\;, \qquad t_a = f_t(q_a)\;,
\end{equation}
and projected into the shared space as:
\begin{equation}
    \hat{t}_n = \mathrm{proj}_{t,n}(t_n)\;, \qquad
    \hat{t}_a = \mathrm{proj}_{t,a}(t_a)\;,
\end{equation}
where $\mathrm{proj}_{t,n}$ and $\mathrm{proj}_{t,a}(\cdot)$ are learnable linear projection layers.
These heads are memory efficient and introduce only a small number of trainable parameters, thereby preserving the overall framework's efficiency.

%%%%%%%%%%%%%%%%%%%%
\subsection{Text-Guided Anomaly Scoring}
To measure image-text alignment, we use cosine similarity.
Let $\mathrm{sim}(\cdot,\cdot)$ denote the cosine similarity function.
For each projected patch embedding $\hat{p}_i$, we compute its similarity to the normal and abnormal text embeddings:
\begin{equation}
    s_i^{p} =
    \left[
        \mathrm{sim}(\hat{p}_i, \hat{t}_n), \;
        \mathrm{sim}(\hat{p}_i, \hat{t}_a)
    \right]\;.
\end{equation}
We then convert these two similarity scores into a probability distribution using a softmax function, and define the patch-level anomaly score as the abnormal-class probability:
\begin{equation}
    A_i^{p} = \mathrm{Softmax}(s_i^{p})\;,
\end{equation}
where $\mathrm{Softmax}(\cdot)$ denotes the softmax operation while $A_i^{p}$ is the probability assigned to the abnormal class.

Following prior work \cite{Zhou2023, Qu2025}, to stabilize the shared embedding space and calibrate the normal--abnormal text directions beyond sparse patch-level supervision, we introduce a global-context regularizer based on the projected global token:
\begin{equation}
     s^{c} =
     \left[
         \mathrm{sim}(\hat{c}, \hat{t}_n), \;
         \mathrm{sim}(\hat{c}, \hat{t}_a)
     \right]\;,
\end{equation}
\begin{equation}
     A^{c} = \mathrm{Softmax}(s^{c})\;,
\end{equation}
where $A^{c}$ is the probability assigned to the anomalous class.
By stacking all patch-level anomaly scores, we obtain a coarse anomaly map:
\begin{equation}
    A^{p} \in \mathbb{R}^{H_p \times W_p}\;,
\end{equation}
where $H_p$ and $W_p$ denote the spatial dimensions of the patch grid.
To recover a dense anomaly heatmap at the input resolution, we upsample the coarse map using bilinear interpolation:
\begin{equation}
    \tilde{A}^{p} = \mathrm{Interp}(A^{p})\;,
\end{equation}
where $\tilde{A}^{p} \in \mathbb{R}^{H \times W}$ is the final pixel-level anomaly map.

%%%%%%%%%%%%%%%%%%%%
\subsection{Training Objective}
During the training stage, the pixel-level anomaly map $\tilde{A}^{p}$ is supervised using the ground-truth mask $M \in \{0,1\}^{H\times{}W}$ and the global-context regularizer ${A}^{c}$ with the class label $C \in \{0, 1\}$.
Thus, we optimize the model using a combination of focal loss, Dice loss, and binary cross-entropy loss:
\begin{align}
    \mathcal{L}_{\mathrm{loc}}   &= \mathcal{L}_{\mathrm{focal}}(\tilde{A}^{p}, M) + \mathcal{L}_{\mathrm{dice}}(\tilde{A}^{p}, M)\;, \\
    \mathcal{L}_{\mathrm{cls}}   &= \mathcal{L}_{\mathrm{bce}}(A^{c}, C)\;, \\
    \mathcal{L}_{\mathrm{total}} &= \alpha \mathcal{L}_{\mathrm{loc}} + \beta \mathcal{L}_{\mathrm{cls}}\;,
\end{align}
where $\alpha$ and $\beta$ are balance coefficients.
During training, only lightweight projection heads are optimized, while both the DINOv3 image encoder and the MobileCLIP2 text encoder remain frozen.

%% file: src/4_experiments.tex
\section{Experiments} \label{sec:experiments}

%%%%%%%%%%
\emph{\textbf{Datasets and competitors.}}
We evaluated \ours{} on MVTec AD \cite{Bergmann2019}, BTAD \cite{Hyun2024}, MPDD \cite{Jezek2021} and VisA \cite{Zou2022}, and against six state-of-the-art models (details in \cref{ssec:zsad}).
We train the model on VisA and test it on the other datasets, following the protocol adopted in \cite{Qu2025, Jeong2023, Cao2024}.
Importantly, the object categories in these datasets differ from those in VisA.
In contrast, for evaluation on VisA, the model is trained on MVTec AD.
We also evaluated the performance of \ours{} in two real-world settings: the MMS dataset \cite{Bonazzi2026} and the real production line of the ICE Lab at the University of Verona \cite{Cunico2024}.

%%%%%%%%%%
\emph{\textbf{Training details.}}
We train the model for 100 epochs on a single NVIDIA RTX 3090 GPU using a machine with 64 GB of system RAM.
The training setup uses a batch size of 64 and a learning rate of $1 \times 10^{-4}$.
We employ the AdamW optimizer with a weight decay of $1 \times 10^{-2}$ to improve generalization.
We extract multi-level features from DINOv3, specifically from blocks 3, 5, 7, and 11.
All images have been resized to $512\times{}512$.
Each projection head is a linear layer of 256 neurons.
We set $\alpha=0.9$ and $\beta=0.1$.

%%%%%%%%%%%%%%%%%%%%
\subsection{Quantitative Results}
\begin{table}[t!]
    \centering
    \small
    \caption{Model complexity comparison.
    In green the improvements of \ours{} compared to the average of all the models.}
    \input{tab/1_efficiency_results}
    \label{tab:1_efficiency_results}
    \vspace{-0.5cm}
\end{table}

The main strength of \ours{} lies in its efficiency.
As shown in \cref{tab:1_efficiency_results}, \ours{} achieves the lowest latency, memory usage, and total parameter count among all compared methods.
The values are obtained on the same server used for training, utilizing the code from \cite{Rolih2024}.
Other details, such as the CUDA optimization settings, the warm-up strategy, and the preprocessing, are available in the GitHub repository.
Compared to the slowest method, \ours{} reduces latency by 76.7\%; compared to the most memory-intensive method, it reduces memory consumption by 69.4\%; and compared to the heaviest model, it reduces the total number of parameters by 88.8\%.
Even when compared with strong recent baselines, the gains remain substantial; with respect to AdaCLIP, \ours{} reduces latency by 53.9\%, memory usage by 52.1\%, and total parameters by 77.8\%; compared to AnomalyCLIP, it reduces total parameters by 80.8\%.

\begin{table*}[t!]
    \centering
    \caption{Comparison of anomaly detection methods on industrial datasets in terms of P-AUROC $\uparrow$.}
    \resizebox{\textwidth}{!}{\input{tab/2_pauroc_results}}
    \label{tab:2_pauroc_results}
    \vspace{-0.4cm}
\end{table*}
\cref{tab:2_pauroc_results} reports on the quantitative comparison of \ours{} with recent ZSAD methods.
\ours{} remains close to the strongest prior methods across several datasets, lower by only 1.3\% from the best reported result in VisA and 5.2\% in MVTec AD.
On MPDD, the gap also remains below 10\%.
Averaged across the four datasets, \ours{} retains more than 88\% of the best pixel-wise performance achieved by the compared methods, showing that the proposed lightweight design preserves strong localization capability despite its substantially reduced complexity.

%%%%%%%%%%%%%%%%%%%%
\subsection{Qualitative Results}
\begin{figure}[t!]
    \centering
    \includegraphics[width=0.95\linewidth]{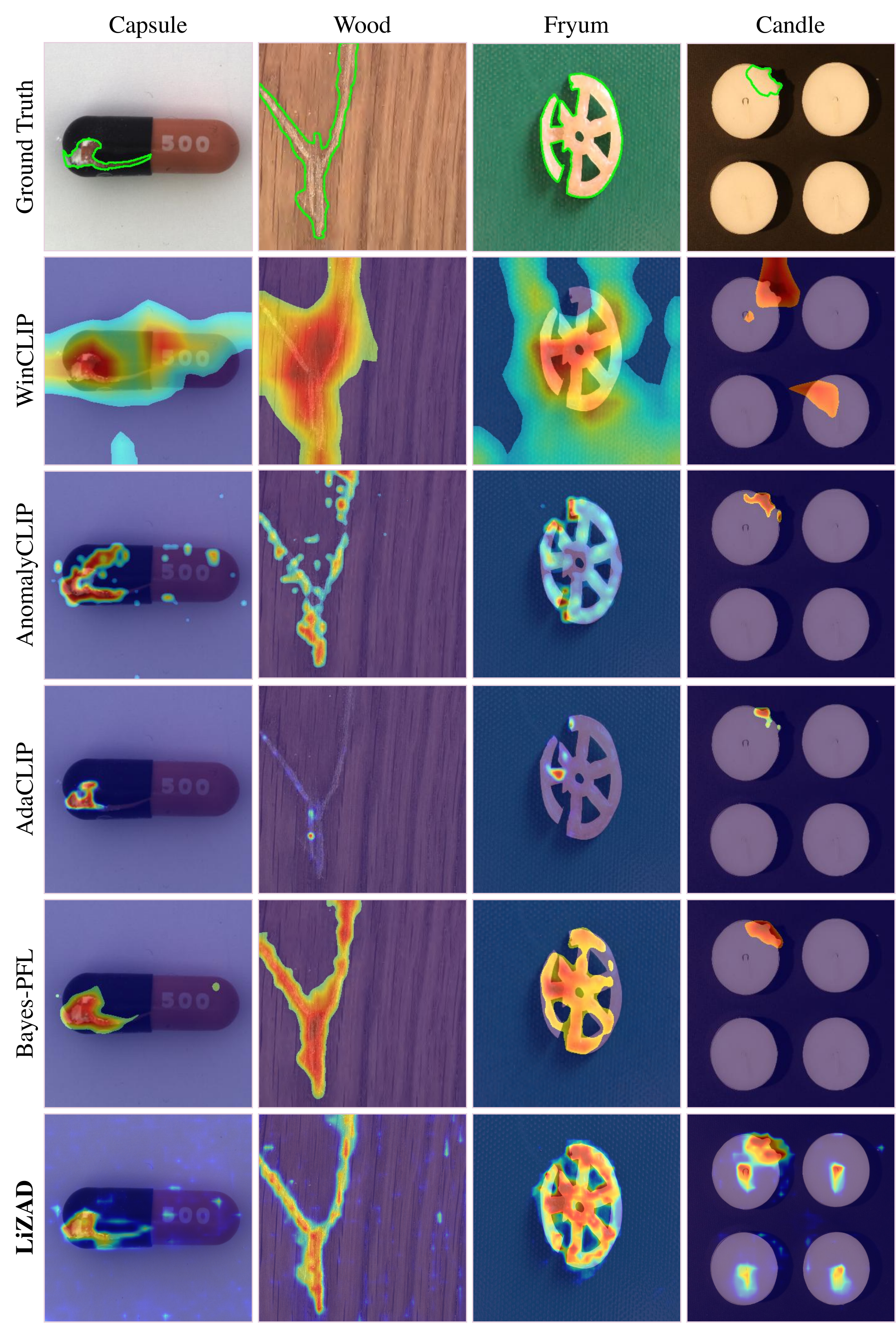}
    \caption{Qualitative comparison of ZSAD methods on different samples from MVTec AD and VisA (extended from \cite{Qu2025}).}
    \label{fig:Qualitative1}
    \vspace{-0.5cm}
\end{figure}
\cref{fig:Qualitative1} presents qualitative results of \ours{} on representative categories from MVTec AD (Capsule and Wood) and VisA (Fryum and Candle).
Across different textures, shapes, and defect patterns, \ours{} produces sharp and concentrated responses, suggesting robust localization behavior in diverse industrial settings.

\begin{figure}
    \centering
    \includegraphics[width=0.95\linewidth]{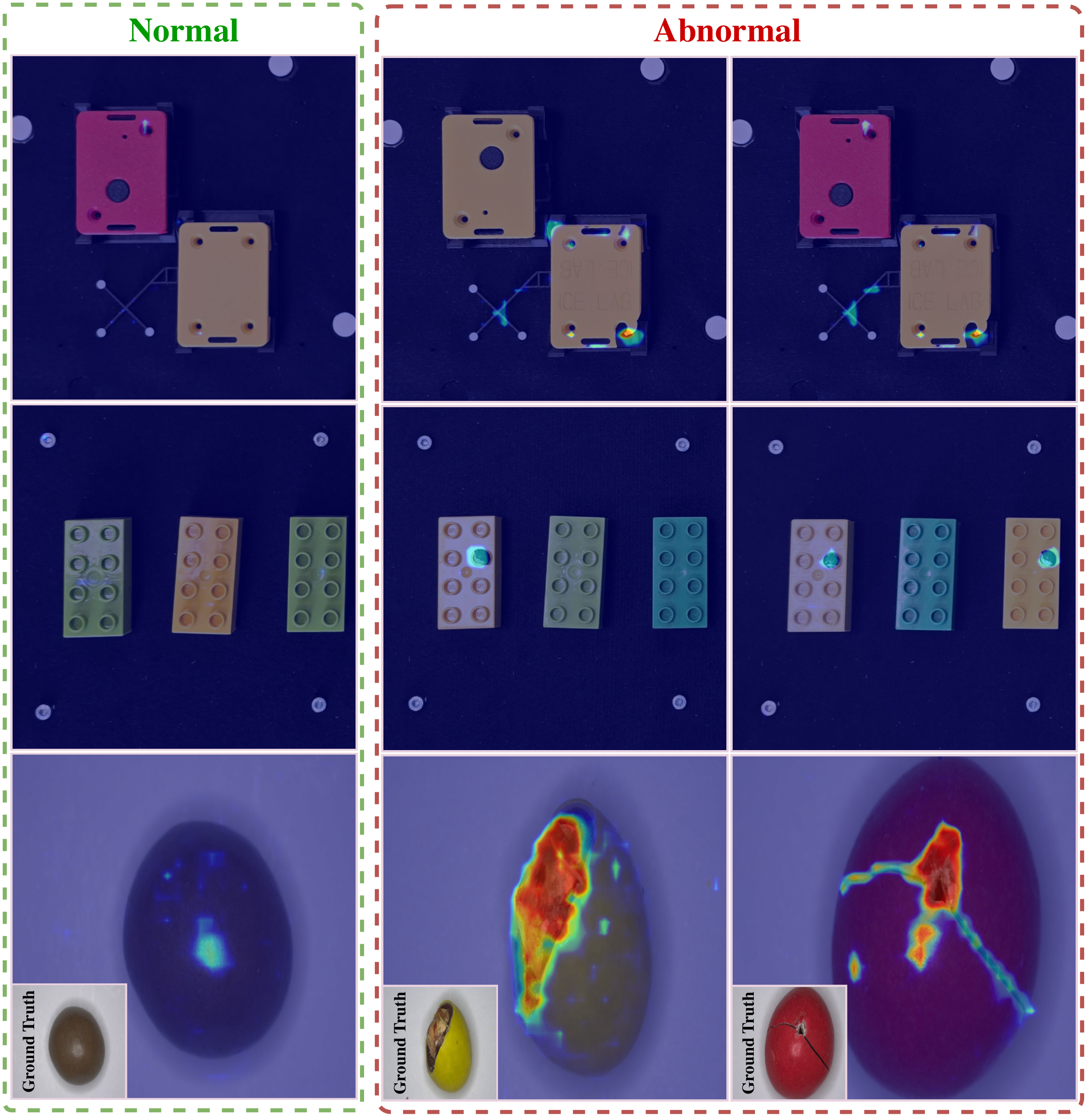}
    \caption{Real-world deployment results of \ours{}.
    The first two rows are images from the ICE Lab at the University of Verona.
    The last row shows images from the MMS dataset.}
    \label{fig:Qualitative2}
    \vspace{-0.5cm}
\end{figure}
\ours{} was successfully deployed on the NVIDIA Jetson NX and Jetson AGX devices and tested on the real production line of the ICE Lab at the University of Verona, achieving an average inference latency of 754.6 ms per sample with a power consumption of 14.8 W on the Jetson NX, while the Jetson AGX delivered a reduced latency of 554.4 ms per sample at a power draw of 28.5 W.
\cref{fig:Qualitative2} illustrates representative examples.
\ours{} was also evaluated on the MMS dataset, which was selected for its highly challenging real-world inspection benchmark.
The last row of \cref{fig:Qualitative2} shows qualitative results.

%% file: tab/1_efficiency_results.tex
\begin{tabular}{lccc}
    \toprule
    \textbf{Model} & \textbf{Latency}           & \textbf{Memory}            & \textbf{Tot. Params.}    \\
                   & \textbf{(ms) $\downarrow$} & \textbf{(GB) $\downarrow$} & \textbf{(M) $\downarrow$} \\
    \midrule
    CLIP \cite{Radford2021}     & \bm3{211.35} & 2.65       & \bm3{427.94} \\
    WinCLIP \cite{Jeong2023}    & 338.56       & 2.74       & \bm2{208.38} \\
    APRILGAN \cite{Chen2023}    & 413.66       & 2.67       & 431.91 \\
    AnomalyCLIP \cite{Zhou2023} & 231.41       & 2.28       & 507.71 \\
    AdaCLIP \cite{Cao2024}      & \bm2{208.86} & \bm2{2.11} & 439.42 \\
    Bayes-PFL \cite{Qu2025}     & 338.50       & 3.30       & 456.72 \\
    \midrule
    \textbf{\ours{}}       & \bm1{96.19}  & \bm1{1.01} & \bm1{97.44} \\
                                & \textbf{\scriptsize{\color{green}{(69.9\%)}}} & \textbf{\scriptsize{\color{green}{(61.5\%)}}} & \textbf{\scriptsize{\color{green}{(74.6\%)}}}\\
    \bottomrule
\end{tabular}

%% file: tab/2_pauroc_results.tex
\begin{tabular}{l|c|c|c|c|c|c|c|c|c}
    \toprule
    \textbf{Dataset} & \textbf{CLIP \cite{Radford2021}}     & \textbf{WinCLIP \cite{Jeong2023}} & \textbf{APRILGAN \cite{Chen2023}}
                     & \textbf{AnomalyCLIP \cite{Zhou2023}} & \textbf{AdaCLIP \cite{Cao2024}}   & \textbf{Bayes-PFL \cite{Qu2025}}
                     & \textbf{\ours{}} \\
                     & ICML'21                              & CVPR'23                           & CVPRW'23
                     & ICLR'24                              & ECCV'24                           & CVPR'25
                     & -- \\
    \midrule
    BTAD     & 30.6 & 32.8 & \bm3{91.1} & \bm2{93.3} & 90.8       & \bm1{93.9} & 83.0 \textbf{\scriptsize{\color{red}{(-10.9\%)}}}\\
    MPDD     & 62.1 & 95.2 & \bm3{94.9} & \bm2{96.2} & \bm1{96.6} & 91.9       & 88.3 \textbf{\scriptsize{\color{red}{(-8.3\%)}}}\\
    MVTec AD & 38.4 & 85.1 & 87.6       & \bm2{91.1} & \bm3{89.9} & \bm1{91.8} & 86.6 \textbf{\scriptsize{\color{red}{(-5.2\%)}}}\\
    VisA     & 46.6 & 79.6 & 94.2       & \bm3{95.4} & \bm2{95.5} & \bm1{95.6} & 94.3 \textbf{\scriptsize{\color{red}{(-1.3\%)}}}\\
    \bottomrule
    \multicolumn{8}{c}{In red, the decrease in percentage of \ours{} with respect to the best models.}
\end{tabular}

%% file: src/5_conclusions.tex
\section{Conclusion} \label{sec:conclusion}

We presented \ours{}, a lightweight ZSAD framework designed for practical industrial deployment.
By combining DINOv3 and MobileCLIP2 with simple projection heads, the proposed architecture achieves competitive anomaly detection while substantially reducing latency, memory usage, and model complexity.
Experiments on public benchmarks and real-world production lines demonstrate that \ours{} offers a favorable trade-off between effectiveness and deployability.